\documentclass[runningheads]{llncs}
\usepackage{graphicx}
\begin{document}
\title{CompareNet: Anatomical Segmentation Network with Deep Non-local Label Fusion}
%
%
\author{
    Yuan Liang \inst{1} \and
    Weinan Song\inst{1} \and
    J.P. Dym \inst{2} \and
    Kun Wang \inst{1} \and 
    Lei He \inst{1}
    }
\institute{
    University of California, Los Angeles, CA 90095, USA \\
    \email{lhe@ee.ucla.edu}
    \and 
    RedNova Health, Irvine, CA 92697, USA
    }
\maketitle              
\begin{abstract}
Label propagation is a popular technique for anatomical segmentation.
In this work, we propose a novel deep framework for label propagation based on non-local label fusion. 
Our framework, named CompareNet, incorporates subnets for both extracting discriminating features, and learning the similarity measure, which lead to accurate segmentation. 
We also introduce the voxel-wise classification as an unary potential to the label fusion function, for alleviating the search failure issue of the existing non-local fusion strategies. 
Moreover, CompareNet is end-to-end trainable, and all the parameters are learnt together for the optimal performance. 
By evaluating CompareNet on two public datasets IBSRv2 and MICCAI 2012 for brain segmentation, we show it outperforms state-of-the-art methods in accuracy, while being robust to pathologies.

\end{abstract}
\section{Introduction}
Deep Convolutional Neural Network (DCNN) models have been widely applied for anatomical segmentation with promising results. 
Most of works formulate the problem as pixel-wise classification, where DCNNs are used to predict the label of a pixel with its surrounding patches.
However, for tasks with a large number of targeting labels and limited labeled data, \textit{e.g.} segmenting brain structures, it is still challenging to design accurate DCNNs \cite{litjens2017survey}. 
Besides classification works, label propagation methods take a different approach.
They first align atlas images with the target one by registrations, and then estimate labels by the label fusion of warped atlases.
By utilizing prior knowledge more directly, they have less dependency on training, and are better constrained with output smoothness.

Label fusion is an important step for label propagation methods. 
Among fusion strategies developed, the non-local label fusion is arguably best.
This strategy labels a pixel by weighted voting of all pixels within a search volume on atlases, with the weights derived from feature-wise similarity of target-voting pixel pairs.
The strategy alleviates registration errors caused by inter-subject anatomical variabilities, leading to accurate delineations \cite{coupe2011patch}. 
Two factors, \textit{i.e.} feature extraction and similarity measure, decide the label fusion accuracy.
For feature extraction, patch intensity was firstly used \cite{coupe2011patch,rousseau2011supervised}, and then hand-crafted features, such as intensity gradients \cite{bai2015multi}, were introduced.
\cite{liao2012sparse} further employs feature selection on candidate features.
For similarity measure, Gaussian kernel is mostly used \cite{coupe2011patch,bai2015multi}.
To reduce errors caused by voting pixels of other anatomies, \cite{tong2015discriminative} uses sparse weights, \cite{wang2012multi} models dependency between voters.
Yet, few has been explored using deep features and trainable similarity measures that can be learnt by optimizing the final segmentation performance.
Moreover, the search failure issue exists for images of high local anatomical variabilities \cite{rousseau2011supervised}, as there can be no pixel of the same anatomy within the search volume after atlas warping.
One work \cite{yang2018neural} learns deep similarity features for label fusion, but it neither explores learnable similarity measures, nor considers the search failure issue.

In this work, we explore a deep non-local fusion framework for automatic anatomical segmentation.
The framework, named \textit{CompareNet}, comprises the following substructures: a \textit{classification subnet} for voxel-wise label prediction, a \textit{features embedding layer} for capturing discriminating features, and a \textit{label fusion subnet} performing label fusion, which employs learnable similarity measures with a multilayer perception (MLP). 
Different from existing works, we introduce the voxel-wise classification as an unary potential into fusion function to alleviate the search failure issue.
Moreover, deep features and similarity measure are learnt that cooperate to optimize segmentation accuracy.
Most importantly, CompareNet is end-to-end trainable, and all parameters are optimized together by back-propagation for accurate segmentation.
We test CompareNet on two brain segmentation datasets IBSRv2 and MICCAI 2012 multi-atlas challenge, with one atlas as prior and pre-aligning by B-spline registrations.
We show it outperforms state-of-the-art methods in Dice score, while being robust to \textit{unseen pathologies}, which are synthetic pathologies introduced only on testing images.

\section{Methodologies}
\subsection{Overview}
\begin{figure}[h]
\includegraphics[width=\textwidth]{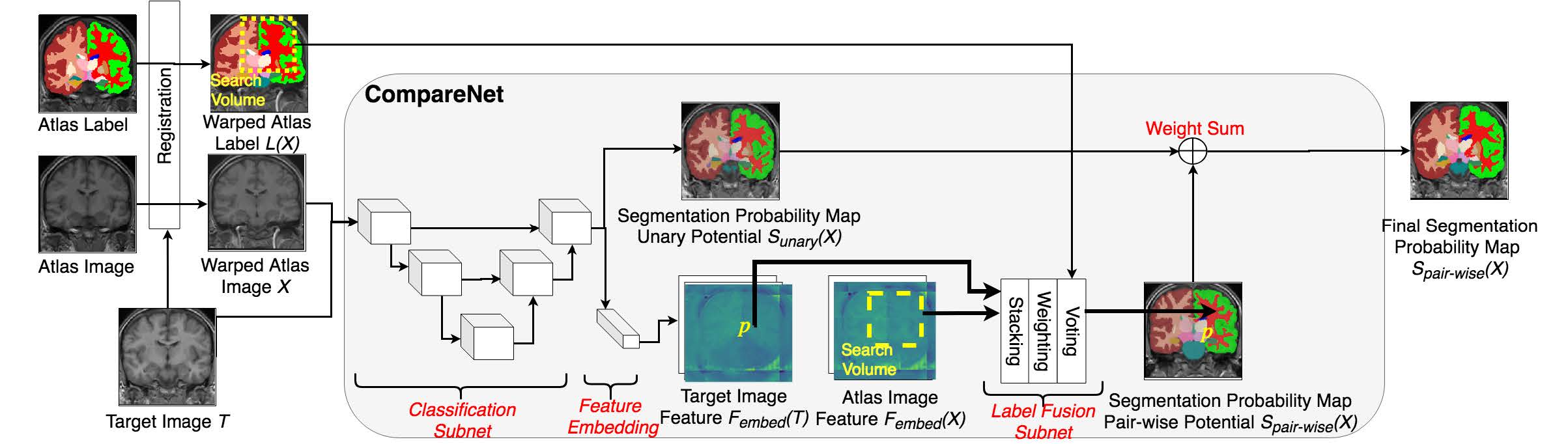}
\caption{Overview of CompareNet. The atlas image and label are aligned to the target image $T$ before CompareNet. A \textit{classification subnet} generates an unary potential of segmentation $S_{unary}(T)$. 
A \textit{feature embedding layer} further embeds deep features of the target and atlas image. 
A \textit{label fusion subnet} takes embedded features to produce a pairwise potential of segmentation $S_{pairwise}(T)$. 
The final segmentation is weighted sum of $S_{unary}(T)$ and $S_{pairwise}(T)$. 
Note all operations in CompareNet are in 3D, while being illustrated as 2D for simplicity.} \label{fig1}
\end{figure}

Fig.~\ref{fig1} shows the overview of CompareNet. 
Given a target image $T$, the atlas image and label are first registered to $T$, resulting in warped atlas image $X$ and label $L(X)$.
Then CompareNet takes $T$, $X$, and $L(X)$ as inputs, and outputs a segmentation probability map $S(T)$ of the target image.
$S(T)$ is a weighted sum of two potentials: an unary potential $S_{unary}(T)$ from a \textit{classification subnet}, and a pairwise potential $S_{pairwise}(T)$ from a \textit{label fusion subnet}. 
The classification subnet predicts voxel-wise label probability of $T$ as semantic segmentation without considering prior knowledge from the atlas.
This aims to alleviate the search failure issue when large anatomical variations exist between atlas and target images such that correct labels do not exist within a local search volume.
Meanwhile, the intermediate feature maps of both $T$ and $X$ from the classification subnet are fed into a \textit{feature embedding layer}, in order to embed discriminative features for label fusion.
The label fusion subnet fuses the atlas labels to generate $S_{pairwise}(T)$, where voxel-wise fusion weights are derived with the embedded features and a MLP as the similarity measure.
Formally, for a target voxel $p$, its unnormalized segmentation probability vector $S_{p}(T)$ is defined as
\begin{equation}
S_{p}(T) = S_{unary,p}(T;\theta_{u})+\alpha S_{pairwise,p}(T;\theta_{p}), \label{eq1} 
\end{equation}
where $S_{unary,p}(T;\theta_{u})$ and $S_{pairwise,p}(T;\theta_{p})$ are unary and pairwise potentials of segmentation, separately.
All probability vectors are in the form $[a_1,...,a_C]$, where $C$ is label classes, and $a_i \geq 0$ represents the probability of class $i$. 
$\theta_u$ and $\theta_p$ are parameters of unary and pairwise potentials, and $\alpha$ weights the two potentials.
CompareNet is end-to-end trainable, and all the parameters ($\theta_u$, $\theta_p$, and $\alpha$) are learnt together for optimal segmentation accuracy.

\subsection{Classification Subnet and Feature Embedding Layer}
\begin{figure}[h]
\centering
\includegraphics[width=0.9\textwidth]{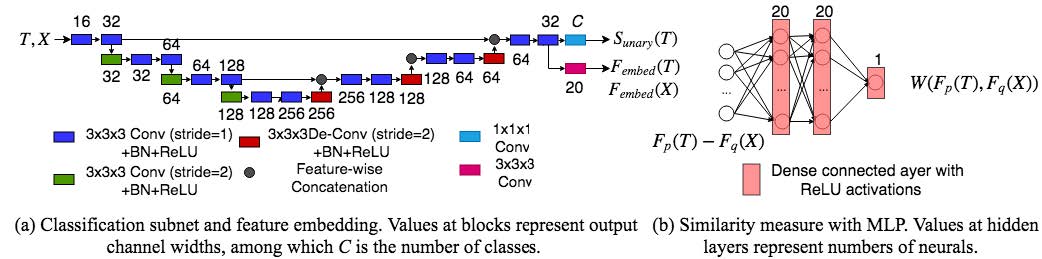}
\caption{(a) Architecture of classification subnet. 
(b) Architecture of the similarity measure with a multilayer perception.} \label{fig2}
\end{figure}

The classification subnet semantically segments the target image, and extracts deep features of both the target image and the atlas image.
The feature embedding layer further embeds the intermediate deep features into a similarity-discriminating space.
Both classification subnet and feature embedding layer are shared by target and atlas images. 
Fig.~\ref{fig2}(a) shows detailed configurations.
The subnet has a encoder-decoder structure with all operations in 3D.
Convolutions and de-convolutions with a stride of 2 are used for down-sampling and up-sampling, and skip-connections are densely employed.
Each convolution is followed by a Rectified linear unit (ReLU) and batch normalization (BN) except the last layer.

We embed deep feature maps before the last layer of the classification subnet with a $1 \times 1 \times 1$ convolution.
This specific layer is chosen because the feature maps at the layer are deep with context information, and have the desired spatial resolution.
The embedded feature maps, denoted as $F_{embed}$, have a fixed feature dimension of 20.

\subsection{Label Fusion Subnet}
The label fusion subnet produces the pairwise potential of segmentation probability map by propagating labels from the warped atlas.
It follows the idea of non-local label fusion methods, where each voxel in the target image is labeled by weighted voting of all voxels within a search volume on the warped atlas image.
The voting weights are determined by similarities between the atlas voxel and the target one, assuming voxels that resemble in features should belong to the same anatomy.
We use the notations in Eq. (\ref{eq1}) and denote the search volume of $p$ as $V_p$. 
The normalized probability vector of $p$ can be written as
\begin{equation}
S_{pairwise,p}(T) = \frac{\sum_{q \in V_p}W(F_p(T),F_q(X))l_q(x)}{\sum_{q \in V_q}W(F_p(T),F_q(X))}, \label{eq2} 
\end{equation} 
where q is a voxel within the $V_p$ on the atlas image. 
$l_q(X)=[0,...,1...,0]$ is the one-hot labeling of length $C$ for $q$, and is 1 at the index of $L_q(X)$. 
$F_p(T)$ and $F_q(X_i)$ are features of the target voxel $p$ and the voxel $q$, separately.
$W(.,.)$ is a similarity measure for two given features, and it is used as the voting weight. 

The feature maps for weighting and their similarity measure \cite{hu2018semi} decide the fusion performance.
We design the feature map to be the feature-wise concatenation of image intensity and its embedded feature map $F_{embed}$.
We employ intensities for low level features as supplement to deep features.
For similarity measure $W(.,.)$, we learn a three layer MLP to capture similarities between any feature vector pair.
The idea is inspired by \cite{santoro2017simple}, where object relations in an image, \textit{e.g. object A is to the left of object B}, are indicated by applying MLPs to take context features of two objects.
Our similarity measure is formulated as
\begin{equation}
W(F_p(T),F_q(X)) = f_\phi(F_p(T)-F_q(x)), \label{eq3} 
\end{equation} 
where $f_\phi$ is the MLP function, and $\phi$ is the set of learnable synaptic weights and bias.
Note that the MLP is shared by all feature pairs.
This aims to encourage greater generalization for computing similarities and avoid it being overfitted to any particular feature pair.
All parameters are learnt by end-to-end training of CompareNet. 
Detailed configuration of the MLP is shown in Fig.~\ref{fig2}(b).

\noindent\textbf{Implementation.} We implement the label fusion subnet for GPU computation by following \cite{yang2018neural}, where weighted voting is decomposed into successive linear operations.
As a result, label fusions of all voxels in an input image patch are parallel, and the subnet can be trained as a part of CompareNet.

\noindent\textbf{Search volume.} A search volume $V_p$ is a $r^3$ cube centered by $p$'s spatially correspondence on the atlas image.
Larger cubes enable more voxels to vote, while costing more computations and memory.
We set $r = 5$ for good segmentation accuracy and moderate GPU memory cost.

\subsection{Atlas Selection and Training}
One atlas image and its expert label from training data are used as priors. 
Despite the limited prior knowledge compared to multi-atlas works \cite{tong2015discriminative,yang2018neural}, CompareNet achieves promising accuracy while saving inference time.
Moreover, the method can be easily extended to multiple atlases by fusing segmentation from each atlas.
To select a proper atlas, we calculate the normalized mutual information (NMI) of all training image pairs.
Following \cite{bai2015multi}, we use the image with the highest NMI within training data as the atlas for both training and testing purposes.

During training, all training images are used as target images $T$.
We train the CompareNet end-to-end by minimizing the generalized Dice loss between the softmax of the final segmentation probability map $S(T)$ and one-hot ground truth $l(T)$, \textit{w.r.t.} $\theta_{u}$, $\theta_{p}$ and $\alpha$. 
Further studies can be carried out by selecting a random image as the atlas in each training step, which generalizes the model to be compatible with any training image as an atlas.
To potentially improve segmentation, one training image that best resembles the target image according to NMI can be selected as the atlas during inference.

\section{Experiments}
\subsection{Experimental Settings}

\noindent\textbf{Dataset.} The CompareNet is evaluated on brain anatomical segmentation on two T1-weighted MR images datasets of IBSRv2\footnote[1]{\url{https://www.nitrc.org/projects/ibsr}} and MICCAI 2012 multi-atlas challenge\footnote[2]{\url{https://my.vanderbilt.edu/masi/workshops/}}.
IBSRv2 has 18 images of $256 \times 256 \times 128$ manually labeled to 32 structures.
We randomly split all images into 13 for training and 5 for testing as \cite{ganaye2018semi} for comparison, and report the average accuracy from three-fold tests.
MICCAI 2012 multi-atlas challenge has 35 T1-weighted MR images around $256 \times 256 \times 260$ that have been labeled to 134 structures.
The dataset was pre-split by the owner into 15 for training and 20 for testing. 
We follow this data split, and report the accuracy on testing images.

\noindent\textbf{Training details.} For B-spline registration and image warping, we use the implementation of SimpleElastix\footnote[3]{\url{https://simpleelastix.github.io/}}.
For training CompareNet, we use mini-batches of 3 $128 \times 128 \times 128$ image patches.
We employ the Adam optimizer with a learning rate of $1 \times 10^{-4}$ and train CompareNet on 3 NVIDIA Titan XP 12GB GPUs.

\begin{figure}[h]
\centering
\includegraphics[width=0.85\textwidth]{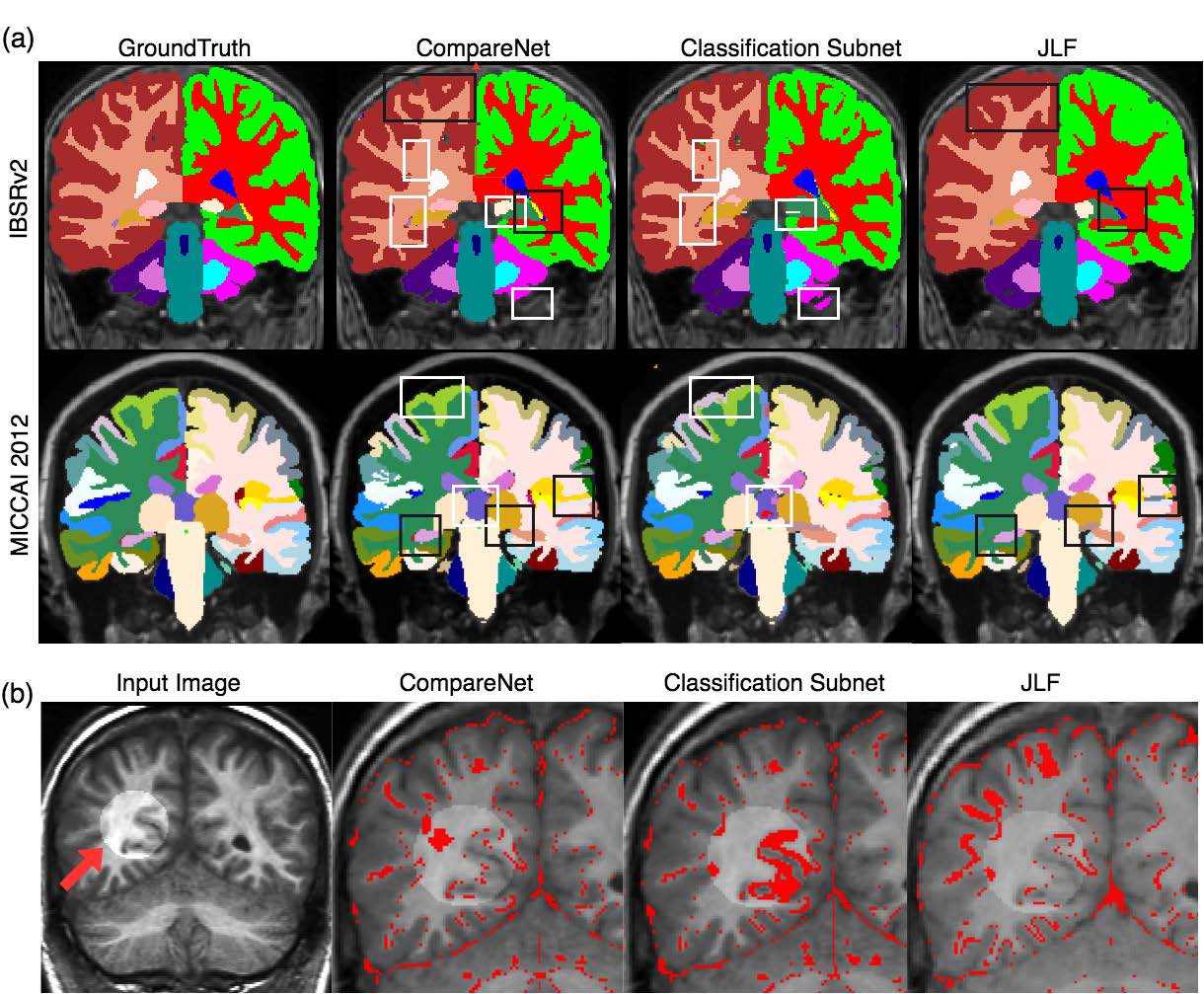}
\caption{(a) Comparison of CompareNet, Classification Subnet, and JLF on two datasets.
White boxes highlight typical areas that are spurious in Classification Subnet while not in CompareNet.
Black boxes highlight typical missing anatomies in JLF while correct in CompareNet.
(b) Segmentation errors of CompareNet, Classification Subnet, and JLF on one IBSRv2 pathological image.
Red arrow shows the pathology on the input image.
Red dots represent label inconsistencies on zoomed segmentations.} \label{fig3}
\end{figure}

\subsection{Results}
\noindent\textbf{Comparison with the state-of-the-arts.}
Table~\ref{tab1} illustrates the comparison of CompareNet with existing works.
Patch-based (PB) \cite{coupe2011patch}, Patch-based with Augmented Features (PBAF) \cite{bai2015multi} and Joint Label Fusion (JLF) \cite{wang2012multi} are label propagation methods, and we use the published implementations\footnote[4]{PB, PBAF: \url{http://wp.doc.ic.ac.uk/wbai/software/}. \\ JLF: \url{http://stnava.github.io/ANTs/}.} for results.
For fair comparison, they take all training images as atlases, and use the same registrations and search volume size as CompareNet.
We determine the rest of their parameters as in \cite{bai2015multi} by using ones that achieve the optimal accuracy on training images, and report results on testing images.
Ganaye \cite{ganaye2018semi}, Brebisson \cite{de2015deep}, and Li \cite{Li2017OnTC} are DCNN based classification methods without label propagation.
We show results of Ganaye and Brebisson from \cite{ganaye2018semi} and \cite{de2015deep} respectively, while producing results of Li with their implementation \cite{Gibson2018NiftyNetAD}.
Note Ganaye (0) represents results of \cite{ganaye2018semi} with the same datasets as CompareNet, while Ganaye (100) uses 100 images elsewhere for training.
The standalone classification subnet of CompareNet, which is also a classification DCNN without propagation, is trained separately for comparison.
We report the average runtime of segmenting one image based on a Intel Core i9-7960X CPU and one NVIDIA Titan XP GPU.

\begin{table}[tp]
\centering
\caption{Comparison of different methods on IBSRv2 and MICCAI 2012 datasets.
Mean Dice scores in percentage are shown in the format of ``mean (standard deviation)''. 
The average runtime for segmenting one image is shown in Minute, and the parameter number for each deep model is shown in Million.
*This method is not directly comparable to CompareNet, because it learns from 100 images elsewhere.}\label{tab1}
\begin{tabular}{l|c|c c|c c}
\hline
 & &\multicolumn{2}{c|}{\textbf{IBSR}} & \multicolumn{2}{c}{\textbf{MICCAI}} \\
\textbf{Method} & \textbf{\# Params} & \textbf{Dice} & \textbf{Runtime} & \textbf{Dice} & \textbf{Runtime} \\
\hline
PB \cite{coupe2011patch} &  N/A & 73.4 (1.8) & 9.8 & 68.8 (3.4) & 11.2\\
PBAF \cite{bai2015multi} &  N/A & 74.2 (5.4) & 10.2 & 67.9 (4.4) & 11.7\\
JLF \cite{wang2012multi} & N/A & 82.4 (0.7) & 41.7 & 73.4 (2.1) & 46.2\\
\hline
Brebisson \cite{de2015deep} & 30.6M & - (-) & - & 72.5 (-) & -\\
Ganaye (0) \cite{ganaye2018semi} & 3.5M & 83.3 (10.0) & - & 73.0 (10.0) & -\\
Ganaye (100)* \cite{ganaye2018semi} & 3.5M & 83.5 (10.0) & - & 73.9 (10.0) & -\\
Li \cite{Li2017OnTC} & 0.9M &81.1 (2.6) & 0.9 & 72.3 (3.2) & 1.0\\
\hline
Classification Subnet & 4.8M & 80.2 (2.6) & 0.6 & 71.9 (4.5) & 1.1 \\
\textbf{CompareNet} & 4.8M & \textbf{84.5 (1.8)} & \textbf{0.7} & \textbf{74.6 (6.4)} & \textbf{1.3}\\
\hline
\end{tabular}
\end{table}

As shown in Table~\ref{tab1}, CompareNet produces the highest Dice scores (84.5\% and 74.6\%) for both datasets.
Compared to label propagation methods (PB, PBAF, JLF), CompareNet benefits from learning optimal features, similarity measures, and fusing with the classification potential.
Moreover, as CompareNet is efficient for GPU computation, it achieves large runtime reduction ($\sim$59x and $\sim$35x compared to JFL) while having higher accuracies.
Compared to classification DCNNs (Brebisson, Ganaye, Li), CompareNet benefits from using the atlas as prior for providing label constraints.
This reduces the mislabeling caused by inter-image local variations, and suppresses spurious segmentation.
It is directly indicated by the Dice improvements (4.3\% and 2.7\%) of CompareNet over the standalone classification subnet, which does not have the label propagation potential.
Moreover, we show CompareNet is more invariant to unseen pathologies than pure classification DCNNs as below.
Fig.~\ref{fig3}(a) visually compares segmentation results of CompareNet, the classification subnet and JLF, where advantages of CompareNet are noticeable within the highlighted areas.

\noindent\textbf{Impact of unseen pathologies.} To evaluate the robustness of CompareNet, we follow the method of \cite{dey2018compnet} by introducing synthetic brain pathologies in testing images.
We design pathologies to be 3D spheres with increased intensities (randomly by 15$\sim$25 in value), and with radius of 10$\sim$20 pixels.
Under one IBSRv2 data split, we keep training data unchanged, and test different methods on testing images with pathologies introduced.
We show mean Dice scores, and their changes from as measured with non-changed testing images in Table~\ref{tab2}.
The results indicate that label propagation methods (PB, JFL) are more invariant to unseen pathologies than classification ones (Li, classification subnet), as anatomical topology is enforced during label propagation.
CompareNet, as a combination of classification and label propagation, shows more robustness than pure classification models (0.6\% mean Dice reduction compared to 1.5\% of the classification subnet).
Fig.~\ref{fig3}(b) shows segmentation errors of different methods on one typical pathological image.

\begin{table}[tp]
\centering
\caption{Comparison of different methods on pathological images.
Mean Dice scores (in \%) and the reductions from those measured with unchanged images are shown in the format of ``mean (reduction)''}\label{tab2}
\begin{tabular}{l|c|c|c|c|c}
\hline
\textbf{Method} & PB \cite{coupe2011patch} & JLF \cite{wang2012multi} & Li \cite{Li2017OnTC} & Classification Subnet & \textbf{CompareNet} \\
\hline
\textbf{Acc.} & 73.0 (0.4) & 82.1 (0.3) & 79.8 (1.3) & 78.7 (1.5) & \textbf{83.9 (0.8)} \\
\hline
\end{tabular}
\end{table}

\subsection{Conclusion}
In this paper, we propose a deep label fusion framework for anatomical segmentation, which combines label propagation and voxel-wise classification. 
Compared to traditional label propagation methods, it learns optimal features and similarity measures for label fusion. 
Compared to existing deep classification methods, it uses training data more directly as priors for labeling. 
We show CompareNet achieves state-of-the-art accuracies on two brain segmentation tasks, and it can also be applied to any anatomical segmentation task without modification. 
We test CompareNet on images with unseen pathologies, and it shows better robustness than pure DCNN classification models.

%
%

\begin{thebibliography}{10}
\providecommand{\url}[1]{\texttt{#1}}
\providecommand{\urlprefix}{URL }
\providecommand{\doi}[1]{https://doi.org/#1}

\bibitem{bai2015multi}
Bai, W., Shi, W., Ledig, C., Rueckert, D.: Multi-atlas segmentation with
  augmented features for cardiac mr images. Medical image analysis
  \textbf{19}(1),  98--109 (2015)

\bibitem{de2015deep}
de~Brebisson, A., Montana, G.: Deep neural networks for anatomical brain
  segmentation. In: Proceedings of the IEEE Conference on Computer Vision and
  Pattern Recognition Workshops. pp. 20--28 (2015)

\bibitem{coupe2011patch}
Coup{\'e}, P., Manj{\'o}n, J.V., Fonov, V., Pruessner, J., Robles, M., Collins,
  D.L.: Patch-based segmentation using expert priors: Application to
  hippocampus and ventricle segmentation. NeuroImage  \textbf{54}(2),  940--954
  (2011)

\bibitem{dey2018compnet}
Dey, R., Hong, Y.: Compnet: Complementary segmentation network for brain mri
  extraction. In: International Conference on Medical Image Computing and
  Computer-Assisted Intervention. pp. 628--636. Springer (2018)

\bibitem{ganaye2018semi}
Ganaye, P.A., Sdika, M., Benoit-Cattin, H.: Semi-supervised learning for
  segmentation under semantic constraint. In: International Conference on
  Medical Image Computing and Computer-Assisted Intervention. pp. 595--602.
  Springer (2018)

\bibitem{Gibson2018NiftyNetAD}
Gibson, E., Li, W., Sudre, C.H., Fidon, L., Shakir, D.I., Wang, G.,
  Eaton-Rosen, Z., Gray, R., Doel, T., Hu, Y., Whyntie, T., Nachev, P., Modat,
  M., Barratt, D.C., Ourselin, S., Cardoso, M.J., Vercauteren, T.: Niftynet: a
  deep-learning platform for medical imaging. In: Computer Methods and Programs
  in Biomedicine (2018)

\bibitem{hu2018semi}
Hu, H., Wang, K., Lv, C., Wu, J., Yang, Z.: Semi-supervised metric
  learning-based anchor graph hashing for large-scale image retrieval. IEEE
  Transactions on Image Processing  \textbf{28}(2),  739--754 (2018)

\bibitem{Li2017OnTC}
Li, W., Wang, G., Fidon, L., Ourselin, S., Cardoso, M.J., Vercauteren, T.: On
  the compactness, efficiency, and representation of 3d convolutional networks:
  Brain parcellation as a pretext task. ArXiv  \textbf{abs/1707.01992} (2017)

\bibitem{liao2012sparse}
Liao, S., Gao, Y., Lian, J., Shen, D.: Sparse patch-based label propagation for
  accurate prostate localization in ct images. IEEE transactions on medical
  imaging  \textbf{32}(2),  419--434 (2012)

\bibitem{litjens2017survey}
Litjens, G., Kooi, T., Bejnordi, B.E., Setio, A.A.A., Ciompi, F., Ghafoorian,
  M., Van Der~Laak, J.A., Van~Ginneken, B., S{\'a}nchez, C.I.: A survey on deep
  learning in medical image analysis. Medical image analysis  \textbf{42},
  60--88 (2017)

\bibitem{rousseau2011supervised}
Rousseau, F., Habas, P.A., Studholme, C.: A supervised patch-based approach for
  human brain labeling. IEEE transactions on medical imaging  \textbf{30}(10),
  1852--1862 (2011)

\bibitem{santoro2017simple}
Santoro, A., Raposo, D., Barrett, D.G., Malinowski, M., Pascanu, R., Battaglia,
  P., Lillicrap, T.: A simple neural network module for relational reasoning.
  In: Advances in neural information processing systems. pp. 4967--4976 (2017)

\bibitem{tong2015discriminative}
Tong, T., Wolz, R., Wang, Z., Gao, Q., Misawa, K., Fujiwara, M., Mori, K.,
  Hajnal, J.V., Rueckert, D.: Discriminative dictionary learning for abdominal
  multi-organ segmentation. Medical image analysis  \textbf{23}(1),  92--104
  (2015)

\bibitem{wang2012multi}
Wang, H., Suh, J.W., Das, S.R., Pluta, J.B., Craige, C., Yushkevich, P.A.:
  Multi-atlas segmentation with joint label fusion. IEEE transactions on
  pattern analysis and machine intelligence  \textbf{35}(3),  611--623 (2012)

\bibitem{yang2018neural}
Yang, H., Sun, J., Li, H., Wang, L., Xu, Z.: Neural multi-atlas label fusion:
  Application to cardiac mr images. Medical image analysis  \textbf{49},
  60--75 (2018)

\end{thebibliography}

\end{document}